\documentclass[journal]{IEEEtran}

\usepackage{CJKutf8}
\usepackage{graphicx}
\usepackage{amsmath}
\usepackage{amsthm}

\begin{document}

\title{On Representation Learning with Feedback}

\author{Hao~Li 
\thanks{H. Li, associate professor and doctoral supervisor, is with Dept. Automation and SPEIT, Shanghai Jiao Tong University (SJTU), Shanghai, 200240, China (e-mail: haoli@sjtu.edu.cn)}}

\maketitle

\begin{abstract}
This note complements the author's recent paper \textit{Robust representation learning with feedback for single image deraining} \cite{Li2021cvpr} by providing heuristically theoretical explanations on the mechanism of representation learning with feedback, namely an essential merit of the works presented in this recent article. This note facilitates understanding of key points in the mechanism of representation learning with feedback.
\end{abstract}


\IEEEpeerreviewmaketitle

\section{Introduction}

An essential merit of the works reported recently in the paper \cite{Li2021cvpr} is the mechanism of representation learning with \textit{feedback}, which is described in Section 3 of the original paper \cite{Li2021cvpr}. Here, we briefly review the mechanism of representation learning with feedback. For expression simplicity, we omit notations of expedient techniques such as scaling images by 0.5 and explicit arguments such as $I_t$ of involved functions in the original paper \cite{Li2021cvpr}.

The \textit{residual map function} $\phi_1$ is obtained via optimization training given in (\ref{eq:residual_map}):
\begin{align}  \label{eq:residual_map}
\min_{\phi_1} ||R_t - \phi_1(.)||
\end{align}
where $R_t$ denotes the residual map truth. The \textit{error detector function} $\varphi$ is obtained via optimization training given in (\ref{eq:error_detector}):  
\begin{align}  \label{eq:error_detector}
\min_{\varphi} ||\frac{\theta_1}{|R_t - \phi_1(.)|} - \varphi(.)||
\end{align}
where $\theta_1$ denotes a tolerance threshold for residual map errors and the residual map error inverses are truncated by 1. The error is computed via (\ref{eq:error_computation}) and feed to $\phi_1$ in a heuristic way as (\ref{eq:error_feed}): 
\begin{align}  \label{eq:error_computation}
err(\varphi) = \frac{\theta_1}{\varphi(.)} - \theta_1
\end{align}
\begin{align}  \label{eq:error_feed}
\phi_1'(.) = \phi_1(.) - err(\varphi) (1 - 2 \phi_1(.))
\end{align}
When the residual map error $|R_t - \phi_1(.)|$ is below the tolerance threshold $\theta_1$, its inverse will be truncated by 1, so $\varphi$ tends to converge to 1 and $err$ tends to converge to 0. This implies that no compensation is made when the residual map error is small enough (or in other words the residual map function is accurate enough), which can maintain stability of training. When the residual map error $|R_t - \phi_1(.)|$ is over the tolerance threshold $\theta_1$, then $err$ tends to converge to $|R_t - \phi_1(.)| - \theta_1$ and compensation is made.

Above equations (\ref{eq:residual_map}), (\ref{eq:error_detector}), (\ref{eq:error_computation}), and (\ref{eq:error_feed}) correspond to equations (2), (3), (4), and (5) respectively in the original paper \cite{Li2021cvpr}. Readers can refer to this original paper for details of these equations in the context of deraining.

In following sections, we provide heuristically theoretical explanations on the mechanism of representation learning with feedback: First, we clarify how the spirit of \textit{feedback} in control theory \cite{Li2024CTPA_SJTU} \cite{Li2024CTPA_Springer} is reflected in the mechanism and explain heuristically why it has the potential to bring performance improvement. Second, we explain from the perspective of \textit{training independence} or \textit{training correlation avoidance} why the error detector function aims at residual map error inverses instead of aiming at residual map errors directly.

\section{Analogue to feedback in control theory}

\begin{figure}[h!]
\begin{center}
\includegraphics[width=0.5\columnwidth]{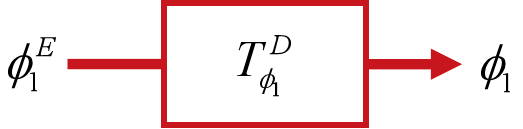}
\end{center}
\caption{Open-loop control process}
\label{fig:open_loop_process}
\end{figure}

The ideal residual map function is denoted as $\phi_1^E$ where the superscript ``E'' means ``expected''. The process of optimization training given in (\ref{eq:residual_map}), which can be regarded as an abstract functional $T_{\phi_1}^D$, can also be regarded as an abstract dynamics block that takes $\phi_1^E$ as input and outputs $\phi_1$. The superscript ``D'' of $T_{\phi_1}^D$ represents certain given datasets for training. By analogue to control theory, this optimization training process can be regarded as an \textit{open-loop control process} illustrated in Fig. \ref{fig:open_loop_process}.

\begin{figure}[h!]
\begin{center}
\includegraphics[width=0.9\columnwidth]{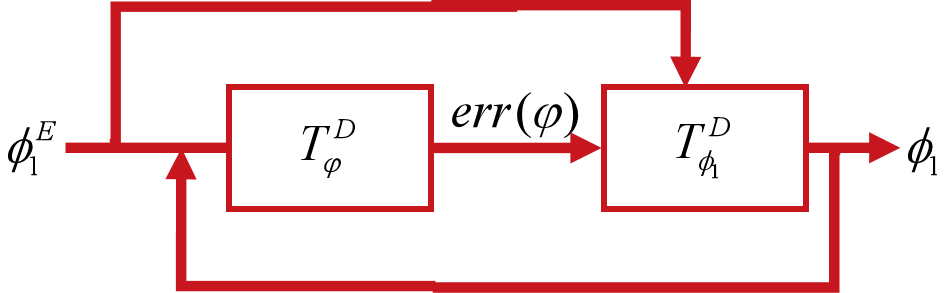}
\end{center}
\caption{Closed-loop feedback control process}
\label{fig:closed_loop_feedback_process}
\end{figure}

Similarly, the process of optimization training given in (\ref{eq:error_detector}) together with (\ref{eq:error_computation}) can be regarded as another abstract functional $T_{\varphi}^D$ as well as an abstract dynamics block. This optimization training process $T_{\varphi}^D$ influences the optimization training process $T_{\phi_1}^D$ indirectly in feedback way. Also by analogue to control theory, such feedback interaction can be regarded as an \textit{closed-loop feedback control process} illustrated in Fig. \ref{fig:closed_loop_feedback_process}.

From control theory we know that if open-loop control does not have desirable control robustness and accuracy, we may resort to closed-loop feedback control for control performance enhancement. Similarly, the architecture with feedback illustrated in Fig. \ref{fig:closed_loop_feedback_process} tends to possess inherent advantage over the open-loop architecture illustrated in Fig. \ref{fig:open_loop_process}. Although it is difficult to strictly prove such advantage, we heuristically borrow block analysis techniques in control theory to provide a heuristically theoretical analysis, which may facilitate understanding of such advantage.

We heuristically treat $T_{\varphi}^D$ and $T_{\phi_1}^D$ as transfer functions that are normally used to characterize dynamics of \textit{linear time-invariant} systems. Here, we heuristically abuse transfer functions. It is expected that the holistic transfer function between the input $\phi_1^E$ and the output $\phi_1$ namely $\phi_1/\phi_1^E$ is the identity transfer function \textbf{1}. For the open-loop architecture illustrated in Fig. \ref{fig:open_loop_process}, the holistic transfer function $\phi_1/\phi_1^E$ is just the open-loop transfer function $T_{\phi_1}^D$ which is usually different from the identity \textbf{1}. It is difficult to obtain a $T_{\phi_1}^D$ being exactly the identity \textbf{1} --- In fact, it is generally difficult to have an open-loop block achieving exactly or even approximating certain expected dynamics.  

On the other hand, for the architecture with feedback illustrated in Fig. \ref{fig:closed_loop_feedback_process}, the holistic transfer function $\phi_1/\phi_1^E$ is computed heuristically as
\begin{align}  \label{eq:feedback_holistic_TF}
\frac{\phi_1}{\phi_1^E} = \frac{T_{\phi_1}^D T_{\varphi}^D}{1 + T_{\phi_1}^D T_{\varphi}^D} (1 + \frac{1}{T_{\varphi}^D})
\end{align}
As mentioned just above, it is generally difficult to have an open-loop block achieving exactly or even approximating certain expected dynamics, so for the open-loop transfer function $T_{\varphi}^D$ as well. However, it would be much easier to obtain an open-loop block achieving dynamics within a flexible range. For example, we may require $T_{\varphi}^D$ to be large enough (analogue to a large proportional factor) or have an accumulating effect of being large (analogue to containing an integral factor $1/s$) without requiring $T_{\varphi}^D$ to exactly achieve certain dynamics. Such conditions which are much easier to satisfy are already sufficient to make the holistic transfer function $\phi_1/\phi_1^E$ given in (\ref{eq:feedback_holistic_TF}) approximate the identity \textbf{1} and hence satisfy our expectation of $\phi_1/\phi_1^E$.

It is worth noting again that above theoretical analysis for the mechanism of representation learning with feedback is only heuristic and by no means holds in strict mathematical sense for representation learning. However, above heuristically theoretical analysis may still help readers understand the inherent advantage of this mechanism.

\section{Training independence}

After clarification of how the spirit of feedback is reflected in the mechanism of representation learning with feedback. Another question arises naturally: To instantiate the spirit of feedback, why does the error detector function $\varphi$ aim at residual map error inverses instead of aiming at residual map errors directly? In other words, why cannot we resort directly to optimization training given in (\ref{eq:error_detector_false}) instead of (\ref{eq:error_detector}) and (\ref{eq:error_computation}) for the sake of residual map error compensation?  
\begin{align}  \label{eq:error_detector_false}
\min_{\varphi} ||(R_t - \phi_1(.)) - \varphi(.)||  
\end{align}

In fact, if optimization training given in (\ref{eq:error_detector_false}) is adopted as $T_{\varphi}^D$, then the abstract dynamics blocks $T_{\varphi}^D$ and $T_{\phi_1}^D$ are highly correlated or coupled because $\varphi$ and $\phi_1$ are trained based on highly correlated data statistics. To facilitate understanding of this point, we may transform (\ref{eq:residual_map}) into (\ref{eq:residual_map2}) which optimizes an incremental term of $\phi_1$ namely $\Delta \phi_1$ within certain sub-manifold of $\phi_1$'s space:
\begin{align}  \label{eq:residual_map2}
\min_{\Delta \phi_1 \in S(\phi_1)} ||(R_t - \phi_1(.)) - \Delta \phi_1(.)||
\end{align}
We can see high correlation between (\ref{eq:error_detector_false}) and (\ref{eq:residual_map2}). Both $\varphi$ and $\phi_1$ are trained based on the same statistics of residual map errors. If optimization of $\phi_1$ via (\ref{eq:residual_map}) is completed, then optimization of $\Delta \phi_1$ via (\ref{eq:residual_map2}) tends to bring no benefit. Consequently, the error detector function $\varphi$, which is trained based on the same statistics of residual map errors via (\ref{eq:error_detector_false}), tends to either bring no benefit or bring the undesirable effect of over-fitting $\phi_1$. This is why we cannot resort directly to optimization training given in (\ref{eq:error_detector_false}).

On the other hand, (\ref{eq:error_detector}) takes advantage of residual map error inverses to heuristically increase independence between data statistics based on which $\varphi$ and $\phi_1$ are trained, namely to increase \textit{training independence} between $T_{\varphi}^D$ and $T_{\phi_1}^D$ or in other words to avoid \textit{training correlation} between $T_{\varphi}^D$ and $T_{\phi_1}^D$. To facilitate understanding of this point, we may give a heuristically theoretical analysis as follows: suppose the residual map error random variable $x$ follows a uniform distribution between $(\theta_1, 1)$ with its probability density function
\begin{align*}
p(x) = \frac{1}{1 - \theta_1} \quad x \in (\theta_1, 1)
\end{align*}
We can derive that its proportional inverse $y \equiv \theta_1/x$ has the following probability density function
\begin{align*}
p(y) = (\frac{\theta_1}{1 - \theta_1})\frac{1}{y^2} \quad y \in (\theta_1, 1)
\end{align*}
so
\begin{align*}
\bar{x} &= \frac{1 + \theta_1}{2}  \\
var(x) &= \frac{(1 - \theta_1)^2}{12} \\
\bar{y} &= \frac{\theta_1}{1 - \theta_1} \ln  \frac{1}{\theta_1} \\
var(y) &= \theta_1 [1 - \theta_1 (\frac{\ln \theta_1}{1 - \theta_1})^2]
\end{align*}
and
\begin{align*}
cov(x,y) &= E[xy] - \bar{x} \bar{y} = \theta_1 [1 + \frac{1 + \theta_1}{2(1 - \theta_1)} \ln \theta_1]
\end{align*}
As $\theta_1 \to 0$, we have
\begin{align}  \label{eq:inverse_independence}
\frac{cov(x,y)}{\sqrt{var(x) var(y)}} \approx \frac{\frac{1}{2} \theta_1 \ln \theta_1}{\sqrt{\theta_1/12}} \to 0
\end{align}
which implies at least quasi-independence between $x$ and its proportional inverse $y \equiv \theta_1/x$. As training independence between $T_{\varphi}^D$ and $T_{\phi_1}^D$ is a key point for the effectiveness of the mechanism of representation learning with feedback, (\ref{eq:inverse_independence}) also accounts heuristically for the effectiveness of (\ref{eq:error_detector}).

By the way, it is worth noting that the heuristic formula (\ref{eq:error_feed}) is designed based on residual map error statistics observed in the context of deraining.

\section{Conclusion}

Heuristically theoretical explanations on the mechanism of representation learning with feedback reported in the author's recent paper \cite{Li2021cvpr} are provided. First, how the spirit of \textit{feedback} in control theory is reflected in the mechanism and why it has the potential to bring performance improvement are explained heuristically. Second, why the error detector function aims at residual map error inverses is explained from the perspective of \textit{training independence} or \textit{training correlation avoidance}. These heuristically theoretical explanations may help readers understand key points in the mechanism of representation learning with feedback.

\ifCLASSOPTIONcaptionsoff
  \newpage
\fi
\bibliographystyle{IEEEtran}
\bibliography{LI_Ref}

\end{document}